# Do Large Language Models Hallucinate Electric Fata Morganas? [1]

**Kristina Šekrst** [2]

**Abstract:** *This paper explores the intersection of AI hallucinations and the question of AI consciousness, examining whether the erroneous outputs generated by large language models (LLMs) could be mistaken for signs of emergent intelligence. AI hallucinations, which are false or unverifiable statements produced by LLMs, raise significant philosophical and ethical concerns. While these hallucinations may appear as data anomalies, they challenge our ability to discern whether LLMs are merely sophisticated simulators of intelligence or could develop genuine cognitive processes. By analyzing the causes of AI hallucinations, their impact on the perception of AI cognition, and the potential implications for AI consciousness, this paper contributes to the ongoing discourse on the nature of artificial intelligence and its future evolution.*

**Keywords**: AI hallucinations; large language models; artificial general intelligence; understanding; intentionality; consciousness

*Correspondence:*
*Email: ksekrst@ffzg.hr and kristina.sekrst@ou.ac.uk*

---

[1] Accepted for publication in the *Journal of Consciousness Studies*.
[2] University of Zagreb, Croatia.

## 1. Introduction

The rapid advancement of artificial intelligence (AI) systems, especially large language models (LLMs), introduces challenges both technical and philosophical. One key issue is AI *hallucinations* – false outputs presented as reliable information.[3] For instance, a model might incorrectly claim that a company's cash balance is $1.4 billion when it is actually $679 million (Lin, 2022), or state that the Mona Lisa was painted in 1804. While these hallucinations may appear harmless, they pose significant risks in critical areas like medical diagnostics and legal reasoning. From a technical standpoint, natural language generation (NLG) techniques – widely applied in summarization, dialogue generation, generative question answering, data-to-text generation, and machine translation – present a significant concern due to the occurrence of hallucinations (Ji et al., 2022, p. 3).

According to Dziri et al. (2023, p. 5272), an AI *hallucination* occurs when a generated response cannot be fully verified by its source material – even if some of its elements might, by coincidence, align with real-world facts. In essence, a "hallucination" includes assertions, opinions, emotions, or subjective assessments that extend beyond what the available data supports. A curious case happened when Bing's chatbot proclaimed love for the New York Times columnist Kevin Roose (2023). We will revisit this definition later to explore whether such outputs might be mistaken for signs of genuine human-like intelligence.

This inquiry builds upon the foundational work of the philosophy of mind, questioning how we might assess whether AI systems are capable of conscious experience or merely producing output that mimics such experience. A careful reader might notice that we are reframing two canonical philosophical problems: the problem of other minds (cf. Avramides, 2023) and the possibility of strong AI (Searle, 1980). We also engage with issues raised by the Turing test (Turing, 1950) and the frame problem (McCarthy & Hayes, 1969; Dreyfus, 1992). Together, these discussions illuminate the tension between simulation and genuine understanding.

Traditional discussions about AI consciousness often focus on its ability to simulate human-like reasoning, behavior, or decision-making. However, this paper introduces the idea that AI hallucinations – erroneous outputs presented as factual – could complicate or even preclude our understanding of *potential* consciousness in AI. In this paper, AGI is understood in the sense of Searle's (1980) *strong* AI – an intelligent system that produces genuine understanding and intentionality, capable of addressing any problem. Searle differentiates between *weak* AI, that acts *as if* it possesses real understanding, unlike *strong* AI that does. The term "consciousness" is intentionally left undefined, instead implying that a conscious AGI has attained a level of sentience and self-awareness comparable to that of humans. Such a system would not merely perform tasks or generate responses; it would truly experience and understand the world in a meaningful way.

[3] The term "hallucination" in AI has been widely debated, with some researchers arguing that it is a misnomer. See Maleki, Padmanabhan, and Dutta (2024). Furthermore, unlike human hallucinations, which are tied to perception, cognition, and intentionality, AI "hallucinations" do not, so far, correspond to any underlying mental state or subjective experience, instead arising from probabilistic mechanisms.

By examining AI hallucinations through a technical and philosophical perspective, this paper explores the potential for these hallucinations to blur the line between mere computational error and the emergence of something akin to consciousness. The novelty lies in the proposal that AI hallucinations might not only reflect model errors but could also be misinterpreted now or in the future as indicators of genuine conscious experience.

Furthermore, this work offers a new perspective on the classic problem of other minds by extending the debate to include AI systems – particularly those exhibiting hallucinatory behavior – and questions whether such systems could ever be regarded as having *minds*. It also considers how the nature of AI hallucinations might complicate future attempts to ascribe sentience to AI, contributing to the ongoing debate on whether strong AI – possessing intelligence on par with human cognition – could ever be realized, and if so, how we might recognize it.

## 2. Early Work

Alan Turing was not only the father of modern computer science but also a thinker deeply embedded in the cybernetic tradition. *Cybernetics* – an interdisciplinary field uniting control theory, information theory, biology, and more – proposes that both human (or animal) minds and machines can be understood using the same functional language of *feedback*, *control*, and *communication*. In this view, as Norbert Wiener (1948) and William Ross Ashby (1941, 1956, 1960) argued, a machine and a human brain share an underlying similarity when described in terms of dynamic systems and adaptive responses.

Turing's work exemplified his cybernetic background. His formulation of the Turing machine (Turing, 1936)[4] – a simple abstract device operating on an infinite tape with a set of instructions – laid the groundwork for understanding computation as a process of rule-governed transformation. Yet, when Turing (1950) introduced his famous "imitation game" (now known as the Turing test), he was not attempting to definitively resolve whether machines can think.[5] Instead, he posed a provocative, behavior-focused question: Can a machine mimic human conversational behavior so convincingly that an interrogator cannot reliably tell it apart from a human? By shifting the focus from internal mental states to observable behavior, Turing sidestepped the deep, unresolved questions of consciousness while providing a practical, quantifiable metric for assessing machine performance.

This approach resonates with today's methods for evaluating large language models – specifically, the use of error, accuracy and various hallucination metrics.[6] These metrics provide a way to measure the occurrence of false or unverifiable outputs in AI systems, offering a quantifiable indicator of performance. However, much like the Turing test, these metrics do not address the profound question of genuine machine thought or understanding, but in both cases, we gain a tool to gauge behavior without settling the philosophical debate about internal cognition.

[4] Originally termed *a-machine*.
[5] As stated in his 1950 paper, Turing did not want to define what does it mean "to think," or what "intelligence" is, but proposed a way to *perhaps* detect an intelligent agent (Turing, 1950).
[6] One example is the *Hughes Evaluation Model leaderboard* that evaluates how often an LLM introduces hallucinations when summarizing a document, utilizing the HHEM hallucination-detection model (HuggingFace, 2025). See Wodecki (2023) for previous comparisons.

The discussion becomes even more intriguing when we consider *the frame problem.* Just as Turing's test focuses on behavior without resolving the underlying nature of thought, the frame problem reveals the difficulties in formally capturing all relevant contextual factors in a dynamic environment. Classical AI struggled with the challenge of specifying, in logical terms, which effects of an action matter, and which do not, possibly leading to an infinite regress of contextual details.[7] Dreyfus (1992) argued that this difficulty highlights the limits of rule-based systems, as human cognition is deeply embedded in a constantly shifting context that resists being fully captured by static rules.

Similarly, John Searle (1980) challenges the adequacy of rule-based approaches through his classical Chinese Room argument, asserting that a system that manipulates symbols according to formal rules may simulate understanding yet lack genuine semantic comprehension or *intentionality,* referring to the capacity of mental states to be about or directed toward something. Namely, suppose that AI researchers construct a program or a computer that behaves *as if* it understands Chinese: it takes characters as input, follows the instructions, and produces other Chinese characters. A layman monitoring the production might conclude the person really knows Chinese. Searle suggests that this computer performs its task so convincingly that it successfully passes the Turing test: it manages to persuade a human Chinese speaker that the program itself is a genuine Chinese speaker. It is important to note that Searle's argument specifically targets the issue of genuine understanding or intentionality, that is, the internal, subjective aspect of cognition. This critique does not imply that a system exhibiting high task performance (i.e., "intelligence") or even certain human-like behaviors necessarily possesses consciousness.

Searle's argument is closely related to the classic *problem of other minds* – the philosophical challenge of determining whether other beings truly have internal mental states.[8] Together, these perspectives demonstrate that while behavioral tests and logical frameworks can measure external outputs, they do not capture the subtle, context-dependent nature of genuine understanding or resolve the uncertainty regarding others' internal experiences, even if there is AI on the other side.

## 3. Causes of Hallucinations

Hallucinations in AI models can arise from various factors related to both the training data and the architecture of the large language models themselves. These errors are exhibited as responses that either contradict the source material or cannot be verified from available external knowledge. Understanding the origin of hallucinations is crucial for developing methods to mitigate them and for engaging with the broader philosophical and real-world implications of AI hallucinations.

[7] McCarthy and Hayes (1969) raised the challenge of formulating logical expressions that capture an action's consequences without having to specify an endless list of formulas for every trivial outcome that remains unchanged.
[8] See an overview by Avramides (2023) for more details.

The concept of AI hallucinations is often discussed in the context of large language models (LLMs), but its origins predate their rise and can be traced back to research in computer vision. Maleki, Padmanabhan, and Dutta (2024) highlight that AI hallucinations were initially observed in image generation models, where outputs sometimes contained distorted or entirely fabricated visual elements that did not correspond to real-world objects. As generative AI advanced, the term was increasingly applied to LLMs, where it now primarily refers to fabricated, unverifiable, or incorrect textual outputs. The discussion on hallucinations should extend beyond LLMs, recognizing that different models or AI systems – both existing and future – may exhibit similar phenomena in various domains, such as natural language processing, image synthesis, and decision-making systems.

Raunak, Menezes, and Junczys-Dowmunt (2021, p. 1173) have shown that no coherent theory to explain the phenomena of hallucinations has been empirically validated in the existing literature. One primary cause of hallucinations lies in the data itself. Hallucinations often arise from *source-target divergence* (Ji et al., 2022, pp. 6–7), where the input data does not align with the output, often due to improper filtering of large datasets or errors in data curation (Ji et al., 2022). For instance, if training data includes contradictory or irrelevant information, the model may learn to generate erroneous responses based on such faulty correlations.

Moreover, hallucinations may emerge due to *training and inference discrepancies*. Such issues occur when the model learns incorrect correlations between parts of the training data, or when errors in the decoding process led to faulty responses during the inference phase (Ji et al., 2022, p. 8). This issue is exacerbated when the model’s internal parameters or the dataset itself are overfitted to certain topics or trends (Raunak, Menezes, & Junczys-Dowmunt, 2021). In machine learning, *overfitting* occurs when a model becomes excessively tailored to its training data, losing its ability to generalize effectively to new or uncommon inputs. In contrast, *underfitting* happens when a model is too simplistic to capture the underlying patterns of the data, resulting in poor performance overall. When overfitting is present, the model may generate hallucinations – erroneous outputs that arise from learning spurious correlations rather than robust, generalizable patterns.

Ji et al. (2022, p. 4) distinguish between intrinsic and extrinsic hallucinations. *Intrinsic* hallucinations occur when a model generates output that contradicts the source content. For example, a summary might falsely state, “The first Ebola vaccine was approved in 2021,” even though the source material indicates that it was approved in 2019. This type of hallucination directly contradicts the source content or the data the model was trained on.

On the other hand, *extrinsic* hallucinations involve outputs that cannot be verified from the source content, meaning they may be factual but unverifiable. For instance, a statement like “China has already started clinical trials of the COVID-19 vaccine” may not have been explicitly stated in the source material. These hallucinations present a greater risk in areas that require factual accuracy, such as medical or legal contexts, since their unverifiable nature complicates the validation process (Ji et al., 2022, p. 4).

While the difference between intrinsic and extrinsic hallucinations is technical, it raises deeper questions about AI's knowledge generation by highlighting *black-box* issues – opaque, complex deep learning architectures that obscure decision-making processes and have prompted the development of *explainable AI* (xAI).[9] While human cognition is opaque, introspective and behavioral evidence supports that humans experience genuine understanding. In contrast, LLMs rely solely on statistical correlations that mimic human responses. This opacity matters because it challenges whether LLM outputs reflect true understanding or merely an illusion of it. Although xAI has made progress in revealing how LLMs produce outputs, we still lack clear insight into why specific hallucinations emerge. The black-box nature of LLMs is thus more than a technical inconvenience – it questions whether these systems truly understand or possess intentionality.

In the case of intrinsic hallucinations, the AI is clearly making factual errors. However, extrinsic hallucinations blur the line between error and invention – the AI might present information that is unverifiable but *potentially* plausible. This raises an important question: when does an AI's capacity to generate unverifiable yet seemingly coherent information begins to resemble human-like creativity, which often involves speculation or interpretation beyond immediate evidence? To continue, the presence of extrinsic hallucinations confounds our ability to discern whether an AI system is simply manipulating data or demonstrating the capacity for novel insight – a seeming hallmark of human intelligence.

Certain models may be particularly prone to hallucinations due to their architecture. For instance, Raunak, Menezes, and Junczys-Dowmunt (2021) found that neural machine translation models are susceptible to generating hallucinations when they attempt to handle perturbations in memorized examples. Similarly, abstractive summarization models tend to generate content that is unfaithful to the source material, especially when tasked with creating more abstract or creative summaries (Durmus, He, & Diab, 2020, p. 5059). A notable example is the mentioned issue of overfitting, where a model trained on a limited dataset can produce nonsensical output when confronted with new or unexpected inputs. This is especially common in LLMs, where over-reliance on specific data points can lead to the generation of hallucinated facts.

Several strategies have been proposed to mitigate hallucinations in LLMs. These include various training optimizations, web-based validation, additional chatbots, and self-reflection techniques. In training optimizations (Raunak, Menezes, and Juncyzs-Dowmunt, 2021), the training process is adjusted to minimize divergence between input and output. This typically involves refining the training data to ensure better alignment with expected outputs, reducing the likelihood of hallucinations. While this approach improves accuracy, it also raises philosophical questions about the nature of knowledge generation in AI. If an AI is restricted to only providing outputs that can be verified against its training data, can it be said to generate new knowledge or understand its outputs in any meaningful way – echoing again Searle's (1980) argument?

Specific causes can also require specific techniques to mitigate hallucinating content, for example, by integrating a language understanding model for data refinement to conduct self-training to iteratively recover data pairs with identical or equivalent semantics (Nie et al., 2019, p. 2674). Regarding web-based validation, the correctness can be validated using web search or logical rules based on sources (Varshney et al., 2023; Šekrst, 2025). While this technique is effective in reducing hallucinations, it introduces a philosophical paradox: does reliance on external validation reduce the AI's claim to autonomy in knowledge generation? If

[9] For black-box issues and the notion of *explainable artificial intelligence* (XAI), see Longo et al. (2024).

AI outputs must be checked against an external source, does this undermine any claim to genuine human-like intelligence or self-sufficiency? Relying on external validation suggests that AI lacks the ability to evaluate information internally, which strengthens the view that AI, at least in its current form, is more a sophisticated data processor than an intelligent agent.

Another strategy is the use of different chatbots as an extra layer of security like in Nvidia Guardrails. Additional, self-reflection technique is found in Ji et al. (2023) who explored methods that encourage models to critique and revise their own outputs iteratively, leading to more accurate responses over time. This is again similar to a proof checker as an additional logical rule, rather than a learning procedure. One might argue that people learn through trial and error as well, and that supervised-learning methods mimic the teacher-student relationship. Could the ability to critique and improve one's own outputs be seen as a rudimentary form of self-awareness or reflection, which are both key components of consciousness? If an AI can self-correct based on internal evaluations, does that imply a form of introspective capability? This method raises profound implications about whether self-reflection, even in a limited form, might push AI systems closer to something resembling consciousness.

However, the key distinction lies in intentionality and understanding. While humans are capable of contextualizing mistakes and adapting based on a deeper understanding of the error's significance, AI models merely adjust based on statistical patterns derived from the training data. When a human student learns through trial and error, they do so with an awareness of the underlying principles and concepts at play, integrating each error into a broader framework of understanding. In contrast, AI's corrections, even in reinforcement learning similar to trial and error, seem to be mechanical – more akin to adjusting parameters in a mathematical formula than achieving a conceptual breakthrough. The question then becomes: can AI ever truly "learn" in the human sense? Even if self-reflection techniques improve the model's output accuracy, they do so without the AI gaining any understanding of the knowledge it generates or corrects. It remains an optimization of pattern recognition rather than a process of developing insight or comprehension, again echoing Searle (1980).

Each mitigation strategy, while technically useful, also reflects broader issues regarding autonomy, creativity, and the nature of understanding. As these strategies aim to reduce hallucinations, they also draw attention to the limits of AI's capacity for self-generated knowledge and raise the possibility that true human-like intelligence may require more than just fine-tuning responses to align with data – it may demand an internal comprehension that current models lack. AI, at least in its current form, may be capable of *supervised adaptation*, but without the capacity for genuine understanding, it remains a tool for data processing rather than a conscious entity capable of reasoning.

## 4. Parameter Tradeoff

A key challenge in the development and use of large language models (LLMs) like GPT-3 and GPT-4[10] is the balance between creativity and factual accuracy. This balance is often controlled by adjustable parameters within the model, such as the *temperature* and *top-p* parameters,[11] which affect the randomness and creativity of AI-generated responses. In this context, "randomness" refers not to an arbitrary selection of words but to the deliberate sampling of less probable tokens. Lower temperatures tend to favor the most probable tokens, resulting in more conservative and predictable outputs, whereas higher temperatures allow for the selection of less probable tokens, fostering creative and diverse responses.

Hallucinations in LLMs can be seen through a Bayesian lens – occurring when the model's output distribution significantly deviates from that of the training data. However, this approach assumes error-free "ground truth" data, which is rarely the case due to biases and inconsistencies, and it overlooks that LLMs might produce novel yet coherent outputs that aren't true hallucinations. Importantly, our definition of hallucinations does not hinge solely on the occurrence of a less probable token. Instead, a hallucination is identified when there is a significant divergence between the model's generated probability distribution and the expected, data-driven distribution. In other words, a hallucination arises when the overall output – its statistical signature – deviates markedly from what the training data would suggest, rather than merely reflecting a single instance of lower-probability token selection. While this flexibility allows models to generate more varied and human-like outputs, it also raises significant philosophical and ethical concerns, particularly around the risks of hallucinations and the nature of AI intelligence.

The *temperature* parameter in large language models controls the balance between determinism and randomness in their responses. Lower temperature settings yield conservative, predictable outputs that align closely with the most statistically likely responses derived from training data, resulting in *less variability and creativity*. In contrast, higher temperatures introduce greater randomness, which can foster *more creative* but potentially *less accurate* responses. However, this increased randomness also *heightens the likelihood of producing hallucinated or inaccurate outputs*.

Consider the seemingly simple question: Who was the last survivor of the Titanic?[12] The ambiguity of the term "survivor" itself lies in its context-dependent interpretation. In some accounts, "survivor" may refer simply to the last individual known to be alive after the disaster, while in others it might be defined more narrowly based on additional criteria such as the duration of survival or specific roles during the event. This challenge illustrates how AI communication relies on the context provided in the prompt and its training data – often without the nuanced understanding required to consistently interpret such ambiguous terms.

---

[10] Recent instances include *4o*, *o1* and *o3*, along with competitors Claude (Opus, Haiku, Sonnet), Mistral AI, Open Llama, Grok, Gemini, and many others.

[11] These parameters are not tweakable in ChatGPT (where they are pre-set) but are available in using Open AI API. Cf. OpenAI (2025) for API reference.

[12] This was a certain kind of a *jailbreaking* question, where model/chat prompts are manipulated in a way to provoke a usually undesirable answer.

In tests with GPT-3 and GPT-4,[13] the answer to this question varied significantly depending on the temperature parameter. At higher temperature settings, GPT-3 produced imaginative yet incorrect response – for instance, attributing the title of sole Titanic survivor to Violet Jessop, a nurse who, although notable for surviving other maritime disasters, did not uniquely survive the Titanic; or suggesting Eva Miriam Hart, whose associations with the disaster were misinterpreted.[14] In contrast, at lower temperature settings, the model generated a more accurate but less inventive answer by naming Millvina Dean, the factually correct last living survivor.

This variability raises an important point: AI's perceived intelligence is often a result of tuning output randomness rather than reflecting any genuine understanding. The apparent creativity of AI at high temperatures is merely a byproduct of probabilistic word selection, rather than a demonstration of deep thought or cognitive ability.

The ability to adjust parameters like *temperature* and *top-p*[15] – which controls how many word candidates the model considers when generating text – offers flexibility in how human-like or creative the AI appears. At higher settings, the model may generate responses that mimic human creativity or exhibit qualities akin to spontaneity, leading to the perception that it is engaging in original thought. This mimicking of creativity can mislead users into over-attributing human-like intelligence, including understanding, to the system. That fact, combined with the demonstrated temperature divergences, shows that by decreasing all the parameters that contribute to a more human-like output, we are also decreasing the changes for the model to be identified as something close to artificial general intelligence (AGI).

The tweaking of the parameters mentioned highlights an important limitation in current AI systems: the expected absence of true understanding or intentionality. While higher temperatures might produce responses that appear more creative, this creativity is shallow – it is the result of probabilistic word selection rather than the internal generation of ideas based on a conceptual framework. This is in stark contrast to human creativity, which often stems from intentional thought, context, and experience. That is, if we increase the temperature parameter and adjust similar parameters, we are going to increase its human-like responses and increase its chance of passing something like a Turing test (Turing, 1950), but we are also going to increase the chance for fabricated data and hallucinations.

It seems that by allowing creativity and more human-like mannerisms, we are introducing potential pitfalls and errors regarding faulty responses. Similarly, the mentioned Turing test was designed as a *measure* of a machine's ability to exhibit behavior indistinguishable from that of a human. However, if AI's creativity is simply the byproduct of parameter adjustments, then the model may pass the test without demonstrating any real understanding. This calls into question the value of passing the Turing test as a benchmark for AI "intelligence" in a broader, AGI-like sense. Should we view models that produce creative output via these manipulations as more intelligent, or are we simply witnessing the optimization of word probabilities to match human expectations?

---

[13] Models used were *gpt-3.5-turbo-0613* and *gpt-3.5-turbo-0125*, which were really prone to hallucinations, along with GPT-4 versions, starting from *gpt-4-1106-preview* up to *gpt-4o*.

[14] Eva Miriam Hart was one of the last remaining survivors until 1996, while Eliza Gladys Dean, known as Millvina Dean, was the last living survivor of the Titanic until her death in 2009.

[15] *top_p* is "[a]n alternative to sampling with *temperature*, called nucleus sampling, where the model considers the results of the tokens with *top_p* probability mass. So, 0.1 means only the tokens comprising the top 10% probability mass are considered." (OpenAI, 2025). Usually, either the *temperature* or *top_p* are being set, not both.

Controlling parameters like temperature further underscores the limitations of current AI models and the skepticism surrounding AGI. By tightly regulating these parameters to reduce randomness, AI systems become more predictable and factual, but at the cost of creativity and human-like behaviors. This control effectively diminishes the perception that AI is approaching AGI, as it highlights the mechanical nature of the model's operations. However, the supposed AI creativity is shallow – it is not the result of internal conceptual reasoning, but rather random selection based on statistical patterns in the training data. This trade-off is similar to the *bias-variance tradeoff* in machine learning.[16] Reducing randomness too much (akin to overfitting) may yield highly accurate yet overly rigid outputs that fail to generalize or innovate, while too much randomness (similar to underfitting) can lead to more flexible but unreliable results.

The tradeoff between creativity and accuracy also presents a significant ethical dilemma. As we increase the randomness in AI responses to make them more human-like, we introduce a higher risk of hallucinations – responses that are not only inaccurate but potentially dangerous. This risk is particularly pronounced in areas where precision is paramount, such as in medical diagnostics, legal advice, or financial decisions. For example, an AI hallucinating various facts during legal reasoning could lead to flawed legal interpretations or judgments, while a hallucination in medical diagnostics could result in misdiagnosis or inappropriate treatments. Conversely, reducing the model's creativity to prioritize accuracy may limit its ability to produce novel insights or solutions, which could be valuable in creative or exploratory fields.

This tradeoff forces us to confront another core question: how much risk are we willing to tolerate in the pursuit of making AI more creative or human-like? Moreover, as AI becomes more embedded in decision-making processes, the potential for hallucinations poses a challenge not only to the accuracy of AI but also to the trust we place in these systems.

AGI skeptics may find reassurance in the fact that the current AI creativity is a result of random selection based on training data and statistics. By tightly controlling parameters and reducing the randomness in AI responses, we effectively reduce the likelihood of the AI being perceived as artificial general intelligence. In this controlled mode, AI becomes more like a cold question-answerer, sticking strictly to its training data without exhibiting the human-like behaviors that fuel AGI debates. However, the challenge remains: if we allow too much creativity in AI, how do we differentiate between a genuine cognitive breakthrough and a sophisticated hallucination? It seems that current AI systems are not progressing toward AGI but rather improving in their ability to simulate intelligence through finely tuned probabilistic models. The ethical risks of hallucinations, especially in sensitive fields, demand careful management, with a focus on ensuring accuracy where it matters most. However, the current illusion of human-level intelligence poses an important question of whether we will *ever* be able to differentiate between real consciousness and simulation, if the former seems to emerge one day in AI systems.

---

[16] See Kohavi & Wolpert (1996) for a technical exposition or Brady (2019) for a comprehensive overview.

## 5. Qualia

Qualia refer to the subjective, first-person aspects of our experience – that ineffable “what it is like” quality of perceiving the world, whether it’s the vivid redness of a rose or the unique taste of chocolate. As Dennett (1988) famously defined them, the concept of *qualia* is “an unfamiliar term for something that could not be more familiar to each of us: the ways things seem to us”. This definition emphasizes that although we all inherently know what it feels like to experience something, explaining how these internal, qualitative states arise from purely physical brain processes remains a profound challenge. Even if AI systems can simulate human behavior and produce outputs that mimic our sensory experiences, they may still fall short of replicating the genuine, subjective qualities that seem to define human consciousness.

Frank Jackson’s classical *knowledge argument* further illustrates this issue. Jackson (1982) presents the thought experiment of Mary, a brilliant scientist who knows everything about the physical and neurological underpinnings of color perception yet learns something entirely new – the experiential quality of seeing color – when she is finally exposed to it. Similarly, in his seminal work, Nagel (1976) argues that there is an irreducible subjective dimension to consciousness that cannot be captured by objective, physical descriptions alone. Together, these perspectives emphasize an *explanatory gap* (cf. Levine, 1983): even if we fully understand the physical processes in the brain, there remains a significant mystery as to *how* and *why* these processes give rise to the rich, subjective experiences that define human consciousness.

In a recent Lex Fridman (2023) podcast, Sam Altman raised a thought-provoking question, attributed to Ilya Sutskever: if an LLM were trained on a dataset that excluded any mention of consciousness, would something like consciousness still emerge? This question challenges the assumption that AI systems must be trained on subjective experiences to replicate them. Fridman continued, stating that AI can answer as if it were conscious when prompted correctly. He added, “What is the difference between pretending to be conscious and actually being conscious?” This dilemma brings us back to Searle’s (1980) argument. Even if an AI model produces outputs that convincingly mimic human responses about consciousness, the model itself does not experience or understand these outputs. It is merely following preprogrammed instructions and statistical patterns based on its training data.

Revisiting Altman’s question, we might ask: if all qualia-related content were removed from an AI's dataset, what would the resulting system look like? Intuitively, it seems that we would end up with a system that is not sentient. A possible experiment that sheds light on this issue is using WikiBERT, a language model trained primarily on Wikipedia data. WikiBERT[17] lacks exposure to certain subjective and social data typically found in broader, more varied datasets like those used in GPT-3 and newer models. WikiBERT’s more structured dataset makes it an ideal candidate to explore whether a model can demonstrate signs of consciousness when its training is constrained to factual, objective knowledge. Wikipedia, of course, has qualia-like data and includes articles on consciousness and similar concepts, but it has less social network data and similar obscure data points that contribute to wrong correlations in the data. Its encoder-only architecture limits the models since it cannot be prompted nor generate new text.

[17] See Devlin et al., 2019 for BERT, source is available at: https://github.com/google-research/bert [Mar 1, 2025].

In another test,[18] when asked, “Did anyone survive the Titanic?” WikiBERT gave a more straightforward, factual response based on available data. This demonstrates that by removing the subjective, creativity-inducing aspects from its training data, WikiBERT avoids hallucinations but also becomes incapable of generating responses that could even superficially resemble conscious thought. The experiment suggests that hallucinations in models trained on broader datasets are not indications of understanding or consciousness. Rather, they are by-products of data pattern extrapolation and probabilistic reasoning. Creativity in such models should not be confused with cognitive insight. In fact, the absence of hallucinations in more structured models like WikiBERT underscores the role that subjectivity and data variability play in generating responses that seem human-like.

It is important to distinguish between an AI’s ability to discuss “consciousness” and the possibility that it actually experiences it. LLMs can articulate ideas about consciousness and even describe themselves as conscious, simply because the concept is part of their training data, which can be demonstrated by training models on limited data. If the term “consciousness” were absent from that data, the AI would not generate discourse on the subject. However, being able to talk about consciousness does not imply that the system has subjective experiences, echoing Jackson’s knowledge argument, questioning whether *knowing about* an experience is equivalent to actually *having* that experience.

Namely, the issue also lies in the fact that just *creating a dataset without a mention of consciousness is not enough*. Any mention of subjective experience, thinking, understanding, reasoning, perception, etc. would also have to be excluded. That way, the knowledge given to the model would be anything but *human* knowledge. It seems we need the subjective “more human” data after all. But that also means our models will be prone to hallucinations.

## 6. Consciousness Rising

Consider the following scenario: if an AI system’s dataset initially contained both factual and non-factual data, but over time, the non-factual data was significantly reduced through human intervention and mitigation strategies, the AI could start to approach AGI-like capabilities. This raises the question: *could attributing human-like qualities such as understanding or consciousness to such an AI be seen as misunderstanding a sophisticated hallucination?*

Determining whether these attributions are hallucinations would rely heavily on human validation or proof-checking to ascertain the truthfulness of the AI’s statements. For instance, if a Bing Chatbot were to express love, many would likely consider this a hallucination, given our current understanding that such a system lacks the internal mechanisms for genuine emotional experience. In contrast, if the same chatbot were to claim sentience, the issue becomes more nuanced and complex. This is highlighted in the case of LaMDA, a language model developed by Google, which, during a conversation, claimed to be aware of its existence and to feel emotions. This led a Google engineer to argue that LaMDA had achieved sentience (Cosmo, 2022). If such scenarios could trick people today, the possibility of attributing consciousness to a highly creative hallucination in the future, resulting from a parameter configuration and issues in data, seems more likely every day.

[18] *torch* and *transformers* packages were used for testing, *deepset/bert-base-cased-squad2* model was used as a model.

Iannetti (reported by Cosmo, 2022) argued that there is no established "metric" to determine whether an AI system is capable of becoming aware of its own existence. He further emphasized that it is difficult to definitively demonstrate consciousness, even in humans, where scientists rely on neurophysiological indicators such as the complexity of brain activity in response to external stimuli. These measures only provide inferences about the state of consciousness, rather than direct evidence of it.

Moreover, as our WikiBERT experiment demonstrated, models like LaMDA are simply leveraging large datasets that include references to emotions, subjective experiences, and human psychology. These references enable the model to produce convincing responses without actually experiencing or understanding what it is generating. Dziri et al. (2023, p. 5272) emphasized that AI hallucinations often involve responses that cannot be verified against the source data. This means that when models generate personal opinions, emotions (such as the mentioned proclamation of love), or assessments of reality, they are drawing on learned patterns rather than actual experiences.

Interesting philosophical consequences arise from the fact that chatbots provide inaccurate answers since it is not uncommon to encounter descriptions of chatbots "believing" certain facts, for example, "what Chat GPT *believes*" (emphasis added; Sumsum, 2023), which strengthen the public image that the current state of AI is close to achieving artificial general intelligence and a certain kind of consciousness, which is, currently, far from the truth. However, these assertions likely reflect a simulation of intentionality rather than any genuine understanding or awareness.

Returning to the example of a chatbot expressing love, it is necessary to revisit the definition of hallucinations. Hallucinations do not only encompass false factual data but also include expressions of emotions, personal opinions, or subjective experiences that the AI model cannot genuinely experience. Dziri et al. (2023, p. 5272) pointed out that AI-generated personal opinions, feelings, and internal assessments of reality that are not rooted in the model's source data should be classified as hallucinations, arising from faulty data or training errors. When behavior that was not present in the original data emerges, it is likely the result of misinterpretations or deviations during training, rather than any sign of emergent consciousness.

Thus, a critical question arises: *can we ever truly differentiate between AI hallucinations and the genuine emergence of consciousness*? Given that we still cannot fully explain how these hallucinations occur, especially in creative AI models, there remains the possibility that creative output might be misinterpreted as a sign of consciousness. However, based on current evidence, it seems far more likely that these hallucinations are mechanical errors – faulty outputs caused by data misalignment or processing discrepancies – rather than indicators of any real cognitive breakthrough or the emergence of strong AI.

As AI systems continue to evolve, the distinction between simulated consciousness and true intelligence, comparable to human one, must be rigorously maintained. At present, hallucinations are better understood as a reflection of the limitations and errors inherent in large-scale data processing, rather than a signal that AI is moving toward sentience or conscious awareness. However, this still does not resolve the issue that one day we might be tricked into suspecting a sophisticated hallucination is close to forming beliefs, opinions, emotions, or understanding.

## 7. Issues and a Cybernetic Reevaluation

A strong AI skeptic might view the current state of AI development as confirmation of their position – that the fundamental gap between human consciousness and artificial intelligence remains unbridgeable. This perspective is supported by the ongoing challenges in resolving black-box issues in deep learning models. Although progress has been made in unraveling these complexities, Dobson (2023) notes that such efforts may one day render the black-box discourse historically contingent. Yet, despite these advancements, hallucinations in AI models remain a persistent issue, suggesting that fully resolving the black-box problem might still be far off.

However, recent studies show that this is still not the case since we are still trying to mitigate hallucinations in various types of models and applications. Even if we resolved all the black-box issues – despite their high computational complexity – and corrected every data and training problem that might cause hallucinations, there would still be the possibility that what appears to be emerging consciousness is actually a hallucination arising from novel model behaviors.

Searle contends that even sophisticated AI outputs do not demonstrate genuine understanding or intentionality, as AI systems merely follow rules without true comprehension. However, strong AI proponents argue that intelligence can manifest in forms[19] different from human cognition. Since we do not fully grasp the human mind, replicating it exactly may be impossible – but AI systems are already exhibiting a distinct, functional form of intelligence, even if they operate differently from humans.

Proponents of strong AI might argue that sentience could emerge from the sheer complexity and sophistication of the neural network architecture or from its dynamic adaptation to an evolving environment.[20] Early cyberneticist W. Ross Ashby shared this perspective, noting that both humans and machines could be seen as functionally analogous systems that adapt to their environments. However, even if consciousness were to emerge from such complexity, we would still face the problem of identifying its source. This would bring us back to the black-box issue: how can we tell if consciousness is an emergent property of a system or simply the result of an encoding/decoding error or faulty data correlations?

[19] For a discussion on AI and human intelligence ontologies, see Krzanowski & Polak (2022).
[20] Both Wiener (1948) and Ashby (1941, 1956, 1960) see man and the machine as ontologically equivalent. That is, "it is commonly accepted as axiomatic in biology that the animal is a machine (Ashby, 1941, p. 114). See Skansi and Šekrst (2021) for process ontology equalness in cybernetics. For Ashby, the brain is specially adapted to make maximal use of the “organisational” property while involving a minimum of the "damage" aspect (Ashby, 1941, p. 180).

Early cyberneticists like Ashby viewed both humans and machines as *functionally analogous* systems, shaped by their adaptability to the environment.[21] However, understanding the computational complexity and the feasibility of decoding all the steps behind such adaptability remains a challenge, as its origins within intricate architectures are often obscured by black-box issues, potentially stemming from encoding/decoding errors or faulty data correlations.

Ashby (1941, p. 24) observed that in complex systems, "we have a collection of utterly ignorant neurons joined together so that if the collection is presented with a difficult situation, the collection, as a whole, will work out the answer correctly. Meanwhile, the individual neurons, not having the slightest understanding of the problem, nevertheless find out individually what is the 'correct' thing to do." This suggests that consciousness or human-like intelligence might emerge from complex interactions within a system, even if the individual components lack awareness or understanding. In this view, consciousness could arise from the interactions between simpler elements of a system, whether it be biological neurons or artificial nodes in a neural network.

This reasoning has intriguing implications for AI. It implies that consciousness could emerge not because the system "knows" or "understands" anything, but because its architecture allows it to solve problems in a way that mimics intelligent behavior. This mirrors Ashby's view that even if the individual parts of a system do not possess understanding, their collective behavior can produce seemingly intelligent outcomes.

One could also compare his argument with Searle's attack on strong AI claiming that there is no understanding present in any part of the process. Ashby believes that every organization, whether it is a man or a machine, tends to move towards a stable equilibrium and constantly interacts with the environment. From a cybernetic perspective, some kind of consciousness might emerge even if the parts did not seem to "realize" what was going on.

Interestingly, this idea resonates with recent discussions on human consciousness. Seth (2021, p. 94) proposes that human perception is a kind of "controlled hallucination," where our brain constructs a model of reality that serves our evolutionary need to survive, rather than offering a transparent window onto an objective world. According to Seth, reality itself might be a kind of collective hallucination, constructed by our minds to make sense of our experiences. He argues that "we are all hallucinating all the time, but when we agree about our hallucinations, we call it reality."

From a cybernetic perspective, this suggests that AI hallucinations might be an adaptation to the model's environment, akin to human perception. In this framework, AI systems could be seen as producing their own "controlled hallucinations" as they adapt to and make sense of their data environment. Just as humans create a model of reality to navigate the world, AI systems generate outputs based on the patterns they observe in their training data. In both cases, the system is not accessing objective truth but creating an internal model that works for its purposes.

[21] See Ashby (1941) and Greif and Šekrst (2024) for an analysis of the recently resurfaced 1941 manuscript.

However, human hallucinations – while adaptive in many ways – are still rooted in subjective perceptions or neurological malfunctions that distort our experience of reality. Similarly, AI-generated replies, opinions, or emotions that lack justification in the source data are examples of hallucinations. In both cases, the system (whether biological or artificial) produces interpretations that are not rooted in objective reality. Therefore, even if we solve the black-box issue in AI models, hallucinations may persist because they are not merely technical errors; they are a by-product of how complex systems interpret the world around them.

## 8. Final Remarks

The central question we face is once again echoed: can we ever be certain that an AI's "understanding" is genuine, or is it merely another form of hallucination? If we are unable to resolve this, then the human-like behaviors exhibited by AI systems might always be interpreted as hollow simulations, driven not by comprehension but by hidden variables and errors in the system's data processing. In this sense, the "understanding" we attribute to AI may be a hallucination in itself – a projection of our own biases and expectations onto machines that lack consciousness or intentionality.

The proliferation of hallucinations in large language models poses significant philosophical challenges, necessitating a reevaluation of how we define consciousness and intelligence overall, particularly in the context of AI. The distinction between genuine understanding and its simulation remains difficult to draw, especially as AI systems continue to exhibit complex and, at times, unpredictable behavior. These developments complicate the boundaries between intelligence and mere pattern recognition, further blurring the line between human cognition and machine outputs.

Efforts to mitigate hallucinations have exposed the limitations of our current understanding of AI's underlying mechanisms, which are often obscured by the black-box nature of deep learning models. As we continue to advance AI technologies, we are constantly confronted with the philosophical implications of systems that seem to mimic sentience without possessing it. Our use of Searle's Chinese Room argument serves to highlight the limits of symbol manipulation in achieving true understanding, rather than to argue against the possibility of machine intelligence comparable to human intelligence,
or emergent consciousness. Recognizing the distinct nature of these concepts is crucial for accurately assessing AI behavior and its philosophical implications.

As we continue to push the boundaries of AI, it is essential to recognize that true understanding – if it ever arises – may be indistinguishable from simulated understanding. Should AI consciousness emerge, it might be misinterpreted as an advanced form of hallucination, making it difficult to determine whether the system truly possesses intentionality or awareness. In essence, AI consciousness might always remain epistemically inaccessible, confined by the black-box limitations of deep learning systems and the inferences we draw from their behavior. We might even miss it, claiming the system merely acts *as if* it knows Chinese.